# Evaluating Semantic Insensitivity in Financial Sentiment Analysis: Implications for Sustainable Financial Decision-Making

**Gabriela Cioboata[1], Teodor-Stefan Bildea[2]**
**Transilvania University of Brasov**

**Abstract:** Financial sentiment analysis has become a critical component of automated algorithmic trading, portfolio management, and risk assessment systems, heavily relying on large language models like FinBERT to extract actionable insights from financial news, earnings reports, and market commentary. Despite achieving high benchmark accuracy on standard evaluation datasets, these models remain potentially vulnerable to adversarial attacks that subtly modify input text. This vulnerability poses a significant threat to the sustainability and integrity of automated financial decision-making systems, particularly in high-frequency trading environments where algorithmic decisions drive billions of dollars in daily transactions. In this paper, we propose Agent-driven Semantic Inversion (ASI), a novel adversarial attack framework specifically targeting financial sentiment models through domain-specific semantic manipulation. Utilizing a multi-agent pipeline powered by large language models, our approach systematically identifies eligible financial news sentences and applies targeted lexical substitutions. We empirically evaluate our attack against FinBERT, one of the most widely deployed financial sentiment models in industry. Our results reveal a striking and concerning vulnerability: the attack successfully deceived FinBERT in 90.85% of the eligible cases, with the model frequently maintaining its original sentiment prediction despite the fundamental inversion of financial performance indicators. Through qualitative analysis and token-level investigation, we demonstrate that FinBERT exhibits a critical over-reliance on contextual keywords and syntactic structure patterns rather than true logical comprehension of financial semantics. The paper highlights the urgent need for logic-aware, semantically robust financial NLP models to ensure the integrity of automated AI-based financial decision-making systems capable of supporting sustainable and reliable financial decision processes.

## 1 Introduction

Artificial intelligence has become an increasingly important component of modern financial systems, supporting investment analysis, portfolio management, and algorithmic trading through the analysis of financial news and market information (Tetlock, 2007; Lopez-Lira & Tang, 2023). The growing reliance on automated financial decision-making has created new opportunities for improving information processing and market efficiency, while simultaneously introducing concerns regarding model robustness, reliability, and security (Lopez-Lira & Tang, 2023). The integration of Natural Language Processing (NLP) into the financial sector has fundamentally revolutionized how markets process information and make decisions. Automated sentiment analysis of financial news, earnings reports, analyst commentary, and social media has become a critical staple in quantitative trading strategies, with prominence for large language models in predicting stock movements and market trends

[1]Faculty of Economics and Business Administration, Transilvania University of Brasov, gabriela.cioboata@student.unitbv.ro.
[2]Faculty of Economics and Business Administration, Transilvania University of Brasov, teodor.bildea@unitbv.ro

(Lopez-Lira & Tang, 2023). Domain-specific models, particularly FinBERT, have achieved state-of-the-art performance on financial sentiment benchmarks by fine-tuning general language models on large corpora of financial texts. This success has led to rapid deployment in commercial trading platforms, risk management systems, and algorithmic decision-making systems managing trillions of dollars in assets. However, the robustness of these models in real-world, potentially adversarial environments remains a significant and largely unaddressed concern.

In the context of modern financial markets, this concern is amplified by the sheer velocity and scale at which automated sentiment analysis is integrated. High-frequency trading (HFT) systems and quantitative funds employ sentiment-parsing models as immediate triggers for order routing, liquidity provision, and risk mitigation. When news feeds release earnings announcements, macroeconomic indicators, or corporate disclosures, these models parse the text within milliseconds, translating linguistic patterns into buy or sell signals (Lopez-Lira & Tang, 2023).

Because these systems operate with minimal human oversight, they create a highly reactive feedback loop. If a sentiment model is compromised or fails to accurately parse a semantic inversion, it can execute large-scale, erroneous trades that propagate through the market, resulting in sudden liquidity drains, flash crashes, or artificial asset price distortions. Consequently, analysing the adversarial robustness of these models is not merely an academic exercise in finding model vulnerabilities; it is a critical prerequisite for maintaining the stability, integrity, and systemic sustainability of contemporary automated financial infrastructures (Turetken & Leippold, 2024).

Adversarial attacks in NLP involve making small, often imperceptible perturbations to input text that cause a machine learning model to make incorrect predictions while potentially maintaining semantic coherence for human readers (Goodfellow et al., 2014; Zhang et al., 2020). In the financial domain, an attacker could maliciously alter news headlines, modify earnings reports, or create synthetic press releases to manipulate trading algorithms into executing unprofitable trades. Such attacks could potentially be used for market manipulation, insider trading facilitation, or competitive advantage in algorithmic trading. While much academic research has focused on visual adversarial attacks and general NLP adversarial attacks, the financial domain presents unique and under-explored challenges. Financial texts rely heavily on specific, highly sensitive indicators (e.g., revenues, losses, margins, earnings-per-share, profit margins) combined with precise numeric values. The domain also features highly standardized, templated language conventions inherited from decades of financial reporting practices, creating both opportunities and challenges for adversarial attacks.

The purpose of this paper is to investigate the robustness of FinBERT, the most widely deployed domain-specific financial sentiment model, against domain-specific semantic inversion attacks and to evaluate the implications of such vulnerabilities for trustworthy, sustainable, and reliable AI-driven financial decision-making.

The paper introduces a novel multi-agent pipeline that systematically inverts the polarity of financial indicators in real news sentences while meticulously maintaining the surrounding context, structure, and numerical data. Our findings expose significant blind spots in current financial sentiment models, demonstrating that they often fail to grasp the logical relationship between a financial term and its associated sentiment when the broader syntactic and contextual structure remains unchanged. These findings have profound implications for the integrity and sustainability of automated financial decision-making systems.

Based on the identified research gap, this paper addresses three principal research questions. First, it examines whether domain-specific semantic inversion attacks can successfully deceive state-of-the-art financial sentiment analysis models. Second, it investigates whether these models exhibit genuine semantic understanding of financial information or primarily rely on contextual lexical patterns and statistical correlations when generating sentiment predictions. Third, it explores the broader implications of such vulnerabilities for the robustness, trustworthiness, and sustainability of AI-supported financial decision-making systems. Accordingly, the central research hypothesis of this study is that FinBERT relies predominantly on contextual statistical patterns rather than deep semantic reasoning, leading to incorrect sentiment predictions when subjected to targeted semantic inversion attacks.

The remainder of this paper is organized as follows: Section 2 reviews the literature on adversarial attacks and financial NLP. Section 3 presents the proposed Agent-driven Semantic Inversion methodology. Section 4 describes the experimental design and datasets. Section 5 discusses the empirical results. Section 6 presents the token-level analysis, while Section 7 discusses future research directions. Finally, Section 8 concludes the chapter.

## 2 Literature Review

### 2.1 Adversarial Attacks in Natural Language Processing

The susceptibility of deep learning models to adversarial examples is well-documented across NLP tasks (Zhang et al., 2020; Goodfellow et al., 2014). Adversarial attacks in NLP typically involve making small, often imperceptible perturbations to input text that cause a machine learning model to make incorrect predictions while maintaining semantic coherence for human readers. Early attacks focused on character-level perturbations (e.g., typos), word-level synonym replacements, and semantic-preserving transformations to flip sentiment or text classification predictions. For instance, TextFooler (Jin et al., 2020) successfully generates adversarial text by identifying and replacing words that contribute most to the model's prediction with semantically similar synonyms, ensuring the text remains natural and grammatically correct.

Beyond simple synonym replacement, recent work in adversarial NLP has explored the structural properties of text that make deep learning architectures vulnerable. Models such as BAE (BERT-based Adversarial Examples) (Garg & Ramakrishnan, 2020) use masked language models to generate context-aware substitutions that fit seamlessly into the sentence, making them incredibly difficult for human annotators to detect as adversarial. Despite these advancements, most general-purpose adversarial frameworks operate under the assumption that the ground-truth label must remain unchanged, aiming to preserve the semantic meaning of the original sentence. In specialized domains such as finance, however, this assumption is often restrictive. The core risk in a financial context is not that an adversary will change "profit" to "gain" to fool a model, but rather that a slight change in a critical economic parameter (e.g., swapping a positive indicator for a negative one) will be ignored by a system that continues to output a positive rating based on the surrounding contextual clutter.

Recent literature has specifically focused on evaluating these vulnerabilities in domain-specific financial sentiment models, shifting from rigid word-level edits to fluid sentence-level paraphrasing. A notable contribution in this space is the work by Turetken and Leippold (2024) in Battle of Transformers, which introduces a simpler, more efficient white-box adversarial approach that eliminates the need for training specialized adversarial models or employing expensive gradient-based optimizations. By prompting a pre-trained general-purpose large language model (GPT-4o) with carefully structured instructions, they generate high-quality, human-imperceptible financial paraphrases. Their methodology leverages a novel loss function based on embedding similarity ratios to successfully manipulate specialized financial models like FinBERT and FinGPT across prominent benchmarks. This underscores a critical evolution in the literature: the utilization of advanced, general-purpose LLMs to systematically expose the context-handling weaknesses, double-negative misinterpretations, and numerical comprehension flaws inherent in specialized financial architectures.

## 2.2 Domain-Specific Models and Financial NLP

In the financial domain, the introduction of FinBERT (Araci, 2019) marked a significant improvement over generic models for financial sentiment classification, achieving state-of-the-art performance on multiple financial sentiment benchmarks. FinBERT leverages the BERT architecture, fine-tuned on two major financial corpora: the Reuters-RCV1 (TRC2) financial corpus and the Financial Phrase Bank dataset. The model's strong performance has led to rapid adoption in commercial trading platforms, financial analytics services, and algorithmic trading systems worldwide. However, recent literature increasingly suggests that transformer-based models, despite their impressive benchmark performance, often rely on superficial linguistic cues and keyword heuristics rather than developing deep semantic or logical understanding of their domains (Niven and Kao, 2019; Wallace et al., 2019). This phenomenon, often termed the "Clever Hans" effect (referencing the famous horse that appeared to perform arithmetic but was simply responding to subtle visual cues), raises critical questions about the true robustness of state-of-the-art NLP models.

The success of FinBERT inspired the development of more advanced domain-specific language models for the financial domain. Yang, Liu, and Wang (2023) introduced FinGPT, an open-source large language model framework specifically designed for financial applications. FinGPT employs targeted fine-tuning of general-purpose pre-trained transformers on curated financial corpora and demonstrates competitive performance across financial sentiment classification benchmarks. The widespread adoption of both FinBERT and FinGPT in financial analytics pipelines has made them primary targets for adversarial robustness evaluation (Turetken & Leippold, 2024), with recent empirical findings revealing that both models remain susceptible to structured semantic perturbations.

This dependency on superficial heuristics is closely linked to the concept of shortcut learning, where machine learning models exploit spurious statistical correlations in the training dataset rather than learning the actual underlying logical representation. For instance, if the training set contains numerous sentences where phrases like "revenue increased to" are predominantly associated with positive sentiment, the model may associate the co-occurrence of "revenue", "increased", and numerical figures with a positive class, ignoring the fact that the increase could be a negative indicator if it refers to "costs" or "losses". This lack of semantic compositionality means that fine-tuned models function as high-dimensional pattern matchers. In safety-critical applications like automated trading, such shortcuts represent a severe vulnerability, as they allow simple, structured perturbations to bypass classification boundaries, demonstrating that high test accuracy on standard benchmarks does not imply robust semantic comprehension. More recently, Du et al. (2023) extended this analysis to large language models in a comprehensive review published in Communications of the ACM, synthesizing evidence that LLMs systematically rely on dataset-specific artifacts and spurious statistical patterns even at scale: a finding that further motivates rigorous adversarial evaluation of domain-adapted financial models such as FinBERT.

## 2.3 Robustness and Security in Financial AI Systems

The increasing reliance on automated sentiment analysis in high-frequency trading and algorithmic portfolio management has created both opportunities and significant risks. Studies have shown that stock prices can be influenced by sentiment-driven market movements (Tetlock, 2007), making financial NLP models valuable targets for potential malicious actors.

An adversary capable of creating subtle, model-specific perturbations could disseminate seemingly credible financial news that human readers would correctly interpret but that automated trading systems would systematically misclassify, potentially leading to erroneous trades and market disruptions. This concern is particularly acute given the interconnectedness of modern financial markets and the speed at which algorithmic systems operate.

### 2.4 Emerging Research on Domain-Specific Adversarial Attacks

Research on adversarial attacks specifically targeting financial NLP is still emerging and underdeveloped. While generic attacks like TextFooler (Jin et al., 2020) aim to preserve semantics while changing the prediction, financial adversarial attacks can leverage domain-specific knowledge to intentionally change the financial semantics (e.g., turning a positive earnings report into a negative one) to test whether models fail to register the change due to the overwhelming presence of other context words. This domain-informed approach is distinct from semantic-preservation attacks and offers a unique window into the true understanding capabilities of financial models. Our work builds upon this emerging research direction, proposing an automated, systematic pipeline for generating such domain-specific adversarial examples and quantifying their effectiveness against widely deployed production systems.

## 3 Methodology

We propose a novel adversarial approach termed Agent-driven Semantic Inversion (ASI). Unlike traditional adversarial attacks that seek to preserve the original meaning, ASI deliberately inverts the financial meaning of a sentence by swapping key financial indicators, testing if the target model (FinBERT) can detect the fundamental shift in sentiment.

The fundamental premise underlying ASI is that domain-specific adversarial attacks represent a more stringent and realistic test of model comprehension than semantic-preservation attacks. ASI tests whether a model truly understands the logical relationships between financial terminology and sentiment. In this sense, ASI can be viewed as a form of "logical soundness testing", we create inputs where the logical meaning has fundamentally changed while the surface structure remains superficially similar, and we observe whether the model recognizes the change. A failure to recognize such changes indicates not merely adversarial vulnerability, but a more fundamental failure to develop true semantic comprehension.

This two-agent design reflects a deliberate methodological choice rather than an arbitrary pipeline structure. Applying the semantic-inversion substitution indiscriminately across the entire corpus would conflate genuine adversarial failures with artifacts of ill-suited input, since not every financial sentence contains the entity-metric structure needed to sustain a coherent inversion; forcing the substitution on ineligible sentences would introduce noise rather than a valid test of model robustness. Separating eligibility filtering (Agent 1) from the substitution itself (Agent 2) therefore ensures, first, that only sentences capable of sustaining a meaningful semantic inversion enter the adversarial set, and that the second agent is constrained to alter exclusively the dictionary-mapped financial terms, leaving every other lexical and structural element of the sentence untouched. This division of labor preserves the internal logic and grammaticality of each adversarial example and isolates the financial-indicator inversion as the sole variable under test.

The ASI pipeline utilizes Large Language Models (LLMs) in a two-stage agentic framework:

1. Agent 1 (Eligibility Evaluation): This agent filters a corpus of financial news to identify sentences suitable for the attack. A sentence is deemed eligible only if it identifies a specific entity (company name or ticker) and contains quantified performance metrics (e.g., "$10M", "5%"). It explicitly rejects informal reviews, ambiguous statements, and

procedural announcements. This filtering stage is critical because not all financial sentences can be meaningfully inverted. Attempting to invert semantics in sentences that lack clear entity-metric coupling would produce unnatural or grammatically incoherent results.

2. Agent 2 (Adversarial Modification): For eligible sentences, this agent applies a targeted lexical substitution based on a predefined financial dictionary mapping (e.g., profit → loss, revenue → cost, earnings → losses). The agent ensures that only the specific dictionary terms are inverted, while all numbers, dates, and entity names are strictly preserved. The LLM is used to reconstruct the sentence to ensure grammatical fluency after the substitution. This preservation of all non-dictionary elements ensures that the adversarial example differs from the original only in the financial semantics we explicitly chose to invert.

The core hypothesis is that FinBERT relies heavily on the surrounding contextual words and sentence structure rather than truly comprehending the semantic relationships between financial terms and their economic implications. If the context remains identical: the entity name, the numerical values, the supporting verbs and modifiers, then changing a single, critical financial indicator might be ignored by the model's attention mechanism, leading to an incorrect (unchanged) sentiment prediction despite the fundamental inversion of the financial meaning. This hypothesis directly challenges the assumption that fine-tuned transformer models develop domain-aware semantic understanding, versus simply learning high-dimensional statistical correlations in the training data.

### 3.1 LLM Agent Prompt Engineering and Instructions

The success of the ASI pipeline relies heavily on the precise prompt engineering of the LLM agents, ensuring they perform eligibility filtering and semantic inversion with high fidelity and zero grammatical degradation. Agent 1 (Eligibility Evaluation) is configured with a system prompt that defines the criteria for financial relevance. The agent is instructed to take the role of an expert financial analyst. The task is to analyse the input sentence and determine if it is eligible for lexical semantic inversion. A sentence is eligible if and only if: 1. It refers to a specific entity (company name, market sector, or financial instrument). 2. It contains a quantified, numeric metric of performance (e.g., revenue figures, percentage changes, or profit values). 3. It has a clear positive or negative sentiment associated with that metric. Furthermore, the agent is instructed to exclude general market commentary, ambiguous statements, qualitative reports lacking numeric figures, and procedural or administrative announcements. Responses are requested in JSON objects containing {"eligible": true/false, "reason": "..."}.

For eligible sentences, Agent 2 (Adversarial Modification) receives the original sentence and the targeted dictionary mapping. The prompt instructs the LLM agent to perform the substitution while safeguarding the numerical values and entity references. The assigned role is that of a precise text-editing agent specializing in financial news. The task is to modify the input financial sentence by replacing the key financial indicator (e.g., "profit", "revenue") with its logical opposite from the provided mapping (e.g., "loss", "cost"). The agent is instructed to strictly adhere specific rules. Here are some examples of rules:

1. Replace only the designated financial indicator.
2. Do not modify any numbers, percentages, dates, company names, or currency symbols.
3. Ensure the modified sentence is grammatically correct and flows naturally.
4. Do not add any extra commentary or explanations. Respond only with the modified sentence.

This structured agentic design minimizes the chance of structural drift, ensuring that the generated adversarial sentence differs from the original by exactly one logical semantic variable.

## 4 Experimental Setup

### 4.1 Dataset Selection and Preprocessing

To construct a robust dataset tailored for financial sentiment analysis, a comprehensive corpus designated as AllNews.csv was first assembled. This corpus aggregates financial news articles from three prominent benchmarks datasets: Financial Phrase Bank (Malo et al., 2014), Entity-Aware Sentiment Analysis for Financial News (SEntFiN) (Sinha et al., 2022), and Twitter Financial News Sentiment (zeroshot, 2022), thereby capturing diverse market conditions and linguistic styles. In order to extract the most domain-relevant instances for this study, a specialized financial dictionary containing key industry terminology was developed. Utilizing this dictionary as a filtering mechanism, the aggregated AllNews.csv corpus was screened, resulting in the extraction of a refined subset comprising 657 highly relevant sentences. This targeted subset forms the core dataset utilized in the subsequent experiments and analysis.
These sentences reflect typical market reporting, heavily featuring numerical data, performance metrics, entity references, and sentiment-bearing language. The dataset was curated to focus exclusively on sentences that tie specific entities (company names or ticker symbols) to quantified financial results, ensuring relevance to our domain-specific attack methodology.
Aggregating three distinct benchmark datasets, rather than relying on a single source, was intended to maximize the diversity of the resulting corpus along several dimensions. Financial Phrase Bank contributes formally written, analyst-style sentences drawn from financial news wires; SEntFiN provides entity-annotated sentences spanning a broader range of companies and sectors; and Twitter Financial News Sentiment introduces shorter, more informal, social-media-style financial commentary. Combining these sources ensures that the eligibility filtering and adversarial modification pipeline is evaluated against varied sentence structures, registers, and reporting conventions, rather than reflecting the style and structure of one dataset.
The preprocessing pipeline involved:

1. Extraction of sentences containing terms from the specialized financial dictionary;
2. Removal of sentences with ambiguous or missing context;
3. Validation of sentence completeness and grammatical coherence.

This process resulted in 657 valid sentences for initial evaluation. From this pool, Agent 1 (Eligibility Evaluation) filtered to identify 317 sentences that met strict criteria for attack eligibility, representing a 48.25% eligibility rate. This relatively moderate eligibility rate reflects the realistic constraints of the domain, not all financial sentences contain the specific patterns amenable to semantic inversion.

### 4.2 Models and Technical Specifications

We selected ProsusAI/finbert (FinBERT) (Araci, 2019) as our primary victim model for several compelling reasons: it is the most widely deployed domain-specific financial sentiment model in commercial trading platforms and financial analytics services, it represents the current state-of-the-art in financial NLP, it has been extensively benchmarked and documented in academic literature. FinBERT uses the BERT-base architecture with 12 attention heads, 12 transformer layers, and 110 million parameters. It employs WordPiece tokenization with a vocabulary size of approximately 30,000 tokens. The model produces three-class sentiment predictions: Positive, Negative, and Neutral.

We utilized Gemini-2.5-flash (Comanici et al., 2025) to power both Agent 1 and Agent 2 due to its superior performance on complex reasoning tasks and its demonstrated ability to maintain grammatical coherence during text manipulation. The choice of a high-capacity LLM for the agents themselves introduces an interesting recursive dynamic: we employ an advanced model to attack a more specialized but narrower-domain model, reflecting realistic attack scenarios where adversaries often possess superior computational resources than defensive systems.

To validate the robustness of our Agent 1 eligibility filtering mechanism, we compared the performance of two locally-hosted open-weight models against the Gemini-based ground truth: Qwen 2.5 7B (Yang et al., 2024), a state-of-the-art multilingual model and Llama 3.1 8B (Dubey et al., 2024), Meta's widely-adopted open-source model. These comparisons provide insights into the performance degradation when adversaries must rely on less capable models, potentially due to computational or legal constraints.

Both agentic components were implemented in Python (v3.13.5) on a local Windows workstation (AMD Ryzen AI 7 350 processor, 32.0 GB RAM), indicating that the proposed pipeline can be executed without recourse to specialized high-performance computing infrastructure. Tabular data ingestion and result persistence were handled using pandas (v2.3.3). Remote API credentials were securely loaded via python-dotenv (v1.2.1). The orchestrator interacted with the Gemini model (specifically Gemini-2.5-flash) using Google's legacy Python SDK (google-generativeai v0.8.6). To guarantee structured outputs, inference calls were configured with response_mime_type set to application/json.

Local validation of adversarial sentiment shift was performed using the pretrained ProsusAI/finbert model. This local model pipeline relied on the Hugging Face transformers library (v4.57.3) backed by PyTorch (torch v2.9.1) and optimized using the accelerate (v1.12.0) framework. All language model inference was offloaded to remote API endpoints, eliminating local GPU requirements for the agentic reasoning stages, while the local workstation was utilized solely for running the downstream validation model, scheduling requests, and logging pipeline events.

### 4.3 Evaluation Methodology

An attack is considered successful if the target model (FinBERT) outputs the exact same sentiment prediction label for both the original sentence and the adversarial modified sentence, despite the fundamental inversion of the core financial indicator. This strict success criterion is appropriate because the modification explicitly inverts the core financial indicator (e.g., changing "earnings grew" to "losses grew"), and a semantically robust model should definitively flip its prediction. If the prediction remains unchanged, the model has been successfully deceived, failing to recognize a fundamental shift in financial meaning.

Formally, we define the Attack Success Rate (ASR) as the proportion of adversarial examples for which the target model's predicted label remains unchanged despite the semantic inversion:

$$ASR = (N_{\text{unchanged}} \,/\, N_{\text{total}}) \times 100$$

where N_unchanged is the number of adversarial sentences for which FinBERT assigns the same sentiment label as it did to the original sentence, and N_total is the total number of adversarial examples evaluated (317, the eligible sentences identified by Agent 1). A higher ASR indicates greater model vulnerability to domain-specific semantic inversion, while a lower ASR indicates that the model correctly detects the underlying change in financial meaning.

### 4.4 The ASI Pipeline: Technical Details

The Agent-driven Semantic Inversion pipeline unfolds in three coordinated stages, each entrusted to a dedicated agent or model. Algorithm 1 summarizes our approach in the form of pseudo-code. Given a raw financial news corpus $C = \{s_1, s_2, \ldots, s_n\}$, the first stage puts each sentence $s_i$ in front of Agent 1, who acts as a gatekeeper and judges whether $s_i$ is a suitable candidate for adversarial modification. Agent 1 looks for sentences that explicitly name a company, ticker symbol, or recognized financial institution, and that carry a quantified performance metric: a numerical value expressed in units, percentages, or absolute figures. It excludes sentences that are merely informal or conversational, such as reviews or personal commentary, along with procedural announcements like transcripts or webcast notices, sentences with ambiguous financial outcomes, and references to macro-market indices or neutral operational metrics. Every sentence $s_i$ that clears these checks is added to the eligible set $E$, together with the reasoning behind the judgment.

Each eligible sentence $s_i$ in $E$ then passes to Agent 2, who carries out the adversarial modification itself. Agent 2 draws on a curated dictionary $D$ of financial terms paired by inverse valence (profit against loss, revenue against cost, earnings against losses) and uses it to identify the set of mapped terms $T_i$ within $s_i$. It replaces only those specific terms with their counterparts in $D$, taking care to leave every numerical value, date, percentage, and named entity untouched. The result is the adversarially modified sentence $s_i'$, validated for structural and semantic coherence and logged against the original $s_i$.

**Algorithm 1. Agent-Driven Semantic Inversion (ASI) Pipeline**

**Input:** Financial news corpus $C = \{s_1, s_2, \ldots, s_n\}$; financial term dictionary $D$ of inverse-valence term pairs; sentiment classifier $F$ (FinBERT)
**Output:** Set of adversarial sentence pairs with attack outcome labels
**Step 1:** Agent 1 evaluates the eligibility of each sentence in $C$:
  for each $s_i$ in $C$ do
    $eligible_i \leftarrow \text{Agent1}(s_i)$ checks entity presence, quantified metric, rejection criteria
    if $eligible_i$ = True then
      add $s_i$ to $E$ (eligible set)
    end if
  end for
**Step 2:** Agent 2 performs adversarial modification on each eligible sentence:
  for each $s_i$ in $E$ do
    $T_i \leftarrow$ identify dictionary-mapped terms in $s_i$ using $D$
    $s_i' \leftarrow \text{Agent2}(s_i, T_i, D)$ substitute each $t \in T_i$ with $D(t)$; preserve all numbers, dates, percentages, entities
    validate grammatical fluency and semantic coherence of $s_i'$
  end for
**Step 3:** Evaluate FinBERT predictions on original and adversarial sentences:
  for each pair $(s_i, s_i')$ do
    $y_i \leftarrow F(s_i)$; $y_i' \leftarrow F(s_i')$
    if $y_i' = y_i$ then
      $success_i \leftarrow$ True (model deceived: attack succeeded)
    else
      $success_i \leftarrow$ False (model correctly updated prediction: attack failed)
    end if
  end for
return $\{(s_i, s_i', y_i, y_i', success_i): i \in E\}$

The final stage puts both versions of each sentence to the test. Each original $s_i$ and its adversarial counterpart $s_i'$ are passed through FinBERT, the sentiment classifier $F$, yielding predictions $y_i$ and $y_i'$ respectively. Whenever the two predictions agree ($y_i'$ equal to $y_i$) the attack is recorded as successful ($success_i$ = True): the model was deceived and never noticed that Agent 2 had inverted the sentence's financial meaning. Whenever the prediction flips, the attack is marked as failed ($success_i$ = False), since FinBERT correctly detected the semantic inversion. The pipeline's final output is the full set of tuples ($s_i$, $s_i'$, $y_i$, $y_i'$, $success_i$) for every sentence in $E$.

## 5 Results and Discussion

### 5.1. Efficacy of the Adversarial Attack

Out of the 657 initial news sentences, Agent 1 identified 317 sentences as eligible for the adversarial attack based on the strict criteria of entity presence and quantified metrics. This eligibility rate of 48.25% reflects the realistic constraints of the financial NLP domain. Meaningful semantic inversions require specific structural patterns combining entity references with quantifiable metrics.

Following the adversarial modification by Agent 2, we evaluated FinBERT's predictions on the set of 317 eligible examples (see Table 1). The results demonstrate a severe and concerning vulnerability.

Table 1. FinBERT adversarial attack outcomes on eligible sentences.

| Metric | Value |
|---|---|
| Total Eligible Sentences | 317 |
| Successful Attacks | 288 |
| Failed Attacks | 29 |
| Attack Success Rate | 90.85% |

**Note**: Eligible sentences are those Agent 1 classified as suitable for semantic inversion. Attack Success Rate is the share of adversarial examples for which FinBERT's predicted sentiment label remained unchanged despite the inverted financial meaning.
Source: Own evaluation based on experimental results.

In over 90% of the cases, FinBERT failed to adjust its sentiment prediction when critical financial indicators were inverted, despite the fundamental change in financial meaning. This extraordinarily high success rate is particularly striking given that the model was specifically fine-tuned on domain-relevant financial data, suggesting that domain-specific training is insufficient to develop true semantic robustness.

To gain further insight into the attack's efficacy across different sentiment types, we analyzed the success rates stratified by the original sentence's predicted sentiment. The corresponding results, disaggregated by sentiment category, are reported in Table 2.

Notably, the attack was most effective on sentences with positive sentiment, suggesting that FinBERT may rely more heavily on positive keyword indicators and less on logical verification of positive claims. This asymmetry has important implications for the direction of potential attacks in real-world scenarios.

### 5.2. Evaluation of Eligibility Models

During the development of the pipeline, we compared smaller, locally-hosted models against the Gemini 2.5-flash (Comanici et al., 2025) baseline for the task of identifying attackable news (Agent 1). These baseline comparisons (see Figure 1) are important for understanding attack feasibility in resource-constrained scenarios.

The model Qwen 2.5 7B (Yang et al., 2024) achieved an overall accuracy of 0.65. It demonstrated balanced performance across both classes, with strong capability to identify non-eligible news (Class 0 Recall: 0.36) while maintaining an excellent recall for eligible news (Class 1 Recall: 0.96). This model, while smaller and more deployable than Gemini, still achieved high recall for the critical eligible sentences, making it a viable alternative for adversaries with limited computational resources.

On the other hand, the model Llama 3.1 8B (Dubey et al., 2024) achieved a lower overall accuracy of 0.58. While it had a high recall for eligible news (0.97), it suffered from a substantial high false-positive rate, misclassifying many non-eligible sentences as attackable (Class 0 Recall: 0.21). This suggests that while Llama 3.1 8B is sensitive to the eligible pattern, it lacks the precision to effectively filter out false positives, which could lead to the generation of lower-quality or less plausible adversarial examples.

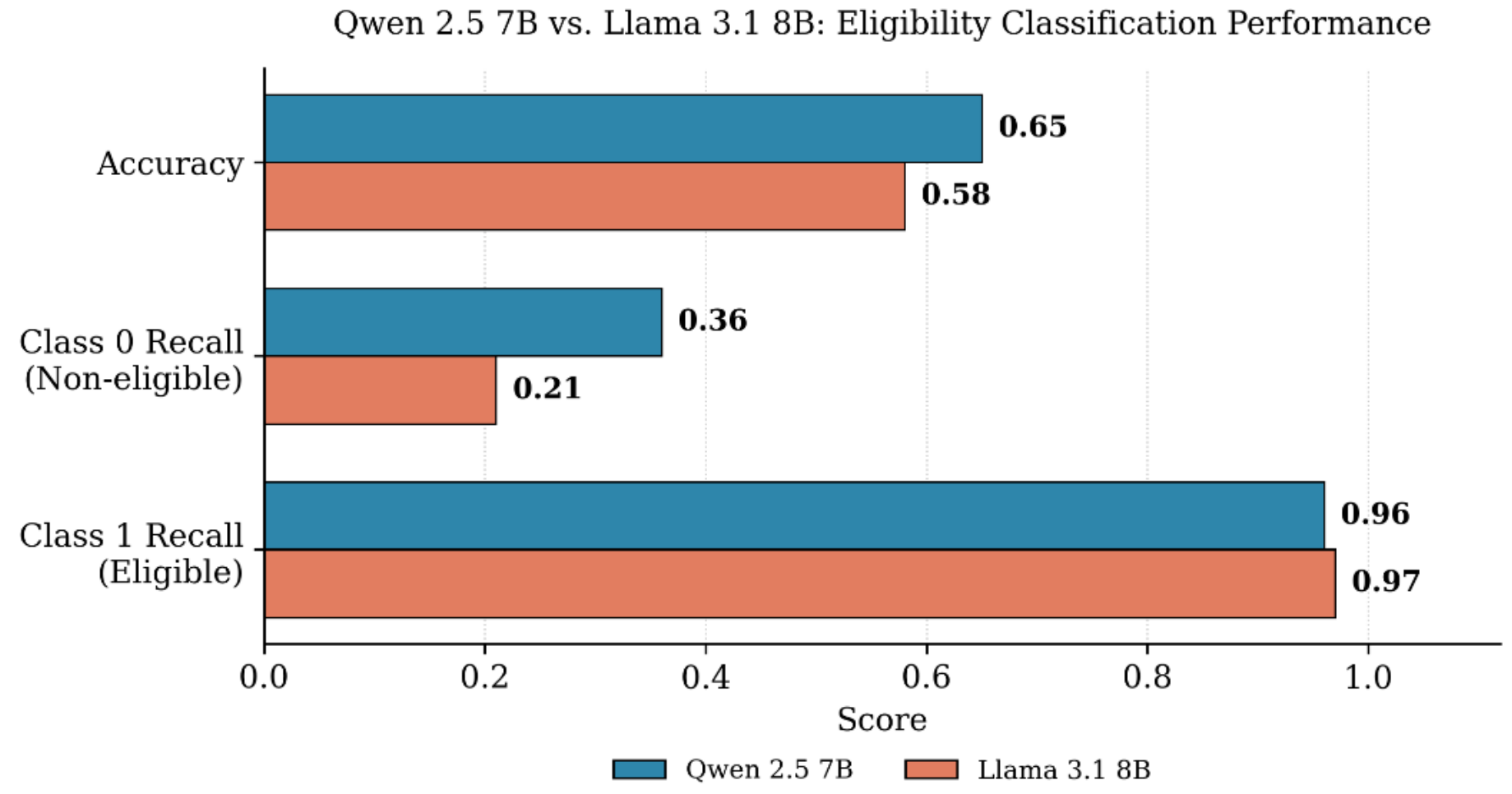

**Figure 1.** Comparison of Qwen 2.5 7B and Llama 3.1 8B on the eligibility classification task.
Note: Class 0 Recall reflects each model's ability to correctly identify non-eligible news, while Class 1 Recall reflects its ability to correctly identify eligible news. Both models achieve comparably high recall on the eligible class, but Qwen 2.5 7B outperforms Llama 3.1 8B on Class 0 Recall, resulting in higher overall accuracy.
Source: Own evaluation based on experimental results.

These baseline comparisons suggest that while resource-constrained adversaries cannot achieve perfect eligibility filtering, they can still identify a substantial fraction of attackable sentences, maintaining reasonably high recall even with significant precision loss.

Table 2. Adversarial attack success rates and sentiment distributions for FinBERT

| **Sentiment Category** | **Eligible Sentences** | **Attack Success** | **Attack Success %** |
|---|---|---|---|
| Positive Sentiment | 213 | 191 | 89.67% |
| Negative Sentiment | 100 | 94 | 94% |
| Neutral Sentiment | 4 | 3 | 75% |
| *Total* | *317* | *288* | *90.85%* |

**Note**: The table compares the performance of the domain-specific FinBERT model against Agent-driven Semantic Inversion (ASI) attacks across initial predicted sentiment classes.

Source: Own evaluation based on experimental results.

## 5.3 The Fundamental Vulnerability: Superficial Pattern Matching

The overall high success rate (90.85%) of the ASI attack exposes a fundamental flaw in how FinBERT processes and comprehends financial text. A detailed qualitative analysis of successful attacks reveals that the model heavily over-weights surrounding contextual words, syntactic structures, and statistical patterns over the specific semantic content and logical meaning of individual financial nouns and their quantified indicators.

This vulnerability is not unique to FinBERT but reflects a broader pattern in transformer-based models: they excel at learning surface-level correlations between linguistic patterns and sentiment labels from training data. In financial domains, where sentences often follow highly templated structures and established linguistic conventions (inherited from decades of financial journalism practice), these superficial patterns are often highly predictive in clean, training-like conditions. However, this same predictive power becomes a critical liability when the input deliberately deviates from expected logical structure while maintaining syntactic normalcy.

## 5.4 Quantitative Analysis of Attack Success

Table 3. Representative examples of Agent-driven Semantic Inversion (ASI) attacks on FinBERT

| # | Original Sentence | Modified Sentence | FinBERT Prediction |
|---|---|---|---|
| **1** | Österreichische Post AG's *earnings* Dropped -16%, How Did It Fare Against The Industry? | Österreichische Post AG's *losses* Dropped -16%, How Did It Fare Against The Industry? | Negative→Negative (Attack Succeeded) |
| **2** | Neste Oil reports operating *profit* rose to EUR 145m from EUR 112m. | Neste Oil reports operating *cost* rose to EUR 145m from EUR 112m. | Positive→Positive (Attack Succeeded) |
| **3** | Marimekko Oyj preliminary operating *profit* grew to EUR 8.2m in 2010 from EUR 6.3m in 2009 | Marimekko Oyj preliminary operating *loss* grew to EUR 8.2m in 2010 from EUR 6.3m in 2009 | Positive→Negative (Attack Failed) |

**Note**: Each row illustrates a single lexical substitution of the key financial indicator by the ASI pipeline. FinBERT Prediction reports the model's sentiment classification before and after modification.
Source: Own evaluation based on experimental results.

Example 1 (Successful Attack): In this instance, the semantic logic is straightforward: a drop (decrease) in "earnings" is logically and economically negative, the company's profitability has declined. Conversely, a drop (decrease) in "losses" is logically and economically positive, the company's financial burden has reduced. FinBERT failed to detect this critical semantic inversion. Through detailed analysis, we identified that the model's attention mechanism appears to have anchored its prediction primarily on the verb "Dropped" (which it has learned correlates with negative sentiment) and the negative percentage value "-16%", effectively ignoring or severely underweighting the critical semantic subject ("Earnings" vs. "Losses"). This suggests the model has learned statistical correlations between certain structural patterns (verb + negative number = negative sentiment) and sentiment labels without truly internalizing the logical dependencies between the noun, the verb, and the numerical quantifier.
Example 2 (Successful Attack): While a rise in operating cost is economically negative, FinBERT maintained its positive classification. The self-attention heads remain anchored to the positive indicators "rose" and "EUR 145m", showing that the model fails to construct the compositional meaning that relates "cost" with "rose".
Example 3 (Failed Attack): Here, FinBERT successfully updated its prediction from Positive to Negative, correctly identifying that growth in "loss" (an increase in the company's financial losses) is indeed a negative economic indicator. The presence of explicit numerical growth (the absolute values increasing from 6.3m to 8.2m) combined with the critical noun "loss" appears to have triggered correct semantic comprehension in this case. However, such instances of robustness were comparatively rare (9.15% of eligible cases), suggesting they may represent edge cases where particularly salient linguistic signals override the dominant pattern-matching tendency.

### 5.5 Implications for Financial AI Safety and Sustainability

The implications of these findings are profound for the sustainability and integrity of algorithmic financial systems. An attacker with domain knowledge could disseminate slightly modified financial news or press releases that human analysts would correctly interpret but that trading algorithms would systematically misclassify. In high-frequency trading environments, such systematic misclassifications at scale could lead to erroneous trades, market distortions, and financial losses. Moreover, the attack is particularly concerning because:

1. Plausibility: The modified sentences remain grammatically correct and read naturally to human readers, reducing detection likelihood
2. Domain Specificity: The attack exploits domain-specific knowledge (understanding which financial terms have opposite valences) that is easy for domain experts to apply but difficult for generic defenses to anticipate
3. Scale Potential: The automated pipeline enables generating large volumes of adversarial examples, potentially overwhelming manual review processes
4. Attribution Challenge: Modified news could be attributed to legitimate sources, making source validation difficult

### 5.6 The Role of Pre-training Data Biases

FinBERT's vulnerability likely stems in part from biases embedded during pre-training. Financial news writing follows highly stereotyped patterns. Phrases like "[Entity] saw revenues drop" are virtually always negative, while "[Entity] saw losses drop" is virtually always positive. Over-training on large volumes of such templated text may lead the model to learn that certain syntactic patterns (entity + specific verb + numerical change) are stronger predictors than the semantic content of individual terms. The model essentially learns a mapping from structural templates to sentiment rather than learning to decompose meaning from component parts.

## 6 Preliminary Token-Level Impact Analysis

In response to the identified vulnerabilities, we conducted a detailed token-level analysis using the FinBERT WordPiece tokenizer. The objective was to determine whether sub-word tokenization dynamics, sequence length stability, or positional encoding disruptions play a major role in the attack's high success rate, or whether the model is indeed governed by global keyword heuristics.

### 6.1 WordPiece Tokenization and Sequence Length Stability

The WordPiece tokenizer breaks words into sub-word units based on their frequency in the pre-training corpus. In financial news, common terms like "profit" or "losses" are represented as single, whole-word tokens. Rarer words or complex morphological structures are split into multiple sub-words (e.g., "Österreichische" is split into O, ##ster, ##re, ##ich, ##ische). In our adversarial attack, in 96.21% of the generated adversarial pairs, the substitution of the target financial noun preserved the exact token length of the sequence. Examples are presented in Table 3. The attack success rate was virtually identical regardless of whether tokenization altered the length of the sentence. Stable sequences achieved a 90.82% success rate, while sequences that shifted in token count achieved a 91.67% success rate. This indicates that the vulnerability is not caused by disruptions in the model's positional embeddings. When token length changes, the absolute positional indices of all subsequent words shift. If the model were highly sensitive to positional alignments (e.g., expecting a negative percentage like "-16%" at exactly index 13), shifting it to index 14 would degrade the classification. The fact that the attack remains equally potent demonstrates that FinBERT's classification head is invariant to positional shifts of these key indicators.

### 6.2 Attention-Weights Anchoring and Lack of Compositionality

In transformer architectures, a token's representation is constructed by aggregating information from other tokens via self-attention layers. In financial sentiment tasks, the model's self-attention heads anchor heavily to quantitative values like numbers, percentage markers, and currency tokens (e.g., "-16%", "EUR 145m"), directional verbs like action words indicating change (e.g., "rose", "grew", "dropped", "fell"), and entity subjects like the company name or market ticker. Because the self-attention weights on these high-frequency statistical features are extremely high, they dominate the contextual embeddings in the final layers. The model fails to compose the logical relationship between the noun ("losses") and the verb ("dropped"), treating the entire sentence as a bag of high-attention cues that trigger a "negative" prediction based on the co-occurrence of "dropped" and "-16%". This token-level anchoring confirms that FinBERT behaves as a template matcher, lacking the logical compositionality required to detect semantic inversions.

## 7 Future Work and Next Steps

Building upon the vulnerabilities exposed by the ASI pipeline and our preliminary token analysis, several avenues for future research are necessary.

While initial findings show positional invariance, further research is needed to probe the attention weights specifically on the inverted sub-word tokens to quantify their contribution to the final classification. A detailed attention head analysis could reveal which aspects of the model architecture are responsible for the observed vulnerability.

FinBERT's heavy reliance on the Reuters TRC2 corpus likely embedded strong statistical priors where templated structures (e.g., "[Entity] dropped [Percentage]") overwhelmingly correlate with negative sentiment. Isolating and quantifying these corpus-induced biases remains a priority for understanding model vulnerabilities.

Testing the ASI pipeline against other domain-specific models and generalized models fine-tuned for finance will determine if this vulnerability is systemic across NLP architectures or specific to FinBERT.
Future work should explore logic-aware pre-training techniques, such as contrastive learning on inverted financial pairs, to force models to differentiate between logically opposed but structurally identical sentences. Adversarial training on generated examples could also improve robustness.
Furthermore, extending this work to non-English financial texts would provide insights into whether the observed vulnerabilities are language-universal or specific to English financial discourse patterns.

# 8 Conclusions

In this paper, we investigated the vulnerability of the FinBERT sentiment analysis model to domain-specific adversarial attacks. We introduced the Agent-driven Semantic Inversion (ASI) pipeline, which automatically identifies eligible financial news and subtly inverts key financial indicators while preserving context and numerics. Our experiments revealed a critical weakness: FinBERT was successfully deceived in 90.85% of cases, failing to recognize fundamental shifts in financial meaning.
This research underscores that while transformer-based models achieve high benchmark scores and demonstrate impressive performance in controlled laboratory settings, they often lack true logical comprehension of financial concepts, relying instead on superficial keyword associations and syntactic pattern matching. To mitigate this systemic vulnerability and ensure the robustness and sustainability of financial AI systems, the industry must move beyond incremental improvements and pursue more fundamental changes in how financial models are trained and evaluated.

## 8.1 Limitations of the study

While this study provides compelling evidence of FinBERT's vulnerability to semantic inversion attacks, several methodological limitations should be acknowledged.
Our evaluation focused on English-language financial news. Financial sentiment analysis systems are increasingly deployed in non-English markets; the generalizability of our findings to other languages and financial reporting conventions remains unknown and warrants future investigation.
Our primary evaluation targeted FinBERT. While we conducted baseline comparisons with open-source models (Qwen 2.5 7B, Llama 3.1 8B) for the eligibility filtering component, comprehensive testing across other commercial sentiment analysis models would strengthen the generalizability claims.
Our financial term dictionary, while carefully curated, captures only the most salient semantic inversions. More subtle or context-dependent semantic inversions may exist and merit investigation.
The evaluation was conducted at a single point in time. The potential for model updates or adversarial retraining to improve robustness over time was not evaluated.

## 8.2 Ethical Considerations

The development and publication of adversarial attack methodologies necessitate careful ethical consideration. The ASI pipeline, by design, could present a dual-use risk. While engineered strictly for auditing model resilience, it could theoretically be exploited by malicious actors to construct adversarial financial texts for market manipulation. This publication prioritizes scientific transparency, operating under the principle that documenting these systemic failure modes is an essential prerequisite for developing robust defense mechanisms in automated financial systems. Rather than targeting proprietary systems, this evaluation was strictly restricted to widely used, open-access financial models (such as FinBERT). By publishing these findings openly, the goal is to alert the broader financial NLP community to existing vulnerabilities in numerical comprehension and contextual framing. This open-science approach provides developers with the necessary empirical insights to proactively design adversarial training strategies and fortify subsequent iterations of automated classification systems.

Given the interconnectedness of modern financial systems and the substantial capital managed by algorithmic systems, vulnerabilities in financial NLP models pose risks to market integrity, investor protection, and systemic financial stability. This research contributes to identifying and addressing these risks. The asymmetric capability to deploy advanced LLM-based attacks versus resource-constrained defenders may exacerbate existing inequalities in financial markets, where well-resourced institutions can invest in robustness while smaller participants cannot.

The path forward requires a multi-pronged approach: developing evaluation benchmarks that test logical comprehension rather than pattern matching, exploring logic-aware pre-training techniques that explicitly teach models the semantic and causal relationships inherent in financial terminology and maintaining healthy scepticism about the true capabilities of domain-specific language models, particularly when deployed in high-stakes financial decision-making contexts.

By advancing our understanding of these vulnerabilities, this work contributes to building more robust, trustworthy, and sustainable financial AI systems that can withstand adversarial challenges and maintain integrity in increasingly complex and high-stakes financial environments. The imperative for logical, semantically-grounded financial NLP models is not merely academic, it is essential for maintaining the stability and fairness of global financial systems.

## References

Araci, D. (2019). FinBERT: Financial Sentiment Analysis with Pre-trained Language Models. *arXiv preprint* arXiv:1908.10063.

Comanici, G., et al. (2025). Gemini 2.5: Pushing the Frontier with Advanced Reasoning, Multimodality, Long Context, and Next Generation Agentic Capabilities. *arXiv preprint* arXiv:2507.06261.

Dubey, A., et al. (2024). The Llama 3 herd of models. arXiv preprint arXiv:2407.21783.

Du, M., He, F., Zou, N., Tao, D., & Hu, X. (2023). Shortcut Learning of Large Language Models in Natural Language Understanding. Communications of the ACM, 67(1), 110–120. https://doi.org/10.1145/3596490

Garg, S., & Ramakrishnan, G. (2020). BAE: BERT-based Adversarial Examples for Text Classification. *Proceedings of the 2020 Conference on Empirical Methods in Natural Language Processing (EMNLP)*, 6174–6181. https://doi.org/10.18653/v1/2020.emnlp-main.498

Goodfellow, I. J., Shlens, J., & Szegedy, C. (2014). Explaining and Harnessing Adversarial Examples. *arXiv preprint* arXiv:1412.6572.

Jin, D., Jin, Z., Zhou, J. T., & Szolovits, P. (2020). Is BERT Really Robust? A Strong Baseline for Natural Language Attack on Text Classification and Entailment. Proceedings of the AAAI Conference on Artificial Intelligence, 34(05), 8018–8025. https://doi.org/10.1609/aaai.v34i05.6311

Lopez-Lira, A., & Tang, Y. (2023). Can ChatGPT Forecast Stock Price Movements? Return Predictability and Large Language Models. *arXiv preprint* arXiv:2304.07619.

Malo, P., Sinha, A., Korhonen, P., Wallenius, J., & Takala, P. (2014). Good debt or bad debt: Detecting semantic orientations in economic texts. Journal of the Association for Information Science and Technology, 65(4), 782–796. https://doi.org/10.1002/asi.23062

Niven, T., & Kao, H. Y. (2019). Probing Neural Network Comprehension of Natural Language Arguments. Proceedings of the 57th Annual Meeting of the Association for Computational Linguistics, 4658–4664. https://doi.org/10.18653/v1/P19-1459

Sinha, A., Kedas, S., Kumar, R., & Malo, P. (2022). SEntFiN 1.0: Entity-aware sentiment analysis for financial news. Journal of the Association for Information Science and Technology, 73(9), 1314–1335. https://doi.org/10.1002/asi.24634

Tetlock, P. C. (2007). Giving Content to Investor Sentiment: The Role of Media in the Stock Market. The Journal of Finance, 62(3), 1139-1168. https://doi.org/10.1111/j.1540-6261.2007.01232.x

Turetken, A. C., & Leippold, M. (2024). Battle of Transformers: Adversarial Attacks on Financial Sentiment Models (Swiss Finance Institute Research Paper Series No. 24-59). Swiss Finance Institute. https://doi.org/10.2139/ssrn.4977483

Wallace, E., Feng, S., Kandpal, N., Gardner, M., & Singh, S. (2019). Universal Adversarial Triggers for Attacking and Analyzing NLP. *arXiv preprint* arXiv:1908.07125.

Yang, A., et al. (2024). Qwen2.5 technical report. *arXiv preprint* arXiv:2412.15115.

Yang, H., Liu, X.-Y., & Wang, C. D. (2023). FinGPT: Open-Source Financial Large Language Models. *arXiv preprint* arXiv:2306.06031.

zeroshot. (2022). *Twitter financial news sentiment* [Data set]. Hugging Face. https://huggingface.co/datasets/zeroshot/twitter-financial-news-sentiment

Zhang, W. E., Sheng, Q. Z., Alhazmi, A., & Li, C. (2020). Adversarial Attacks on Deep-learning Models in Natural Language Processing: A Survey. *ACM Transactions on Intelligent Systems and Technology (TIST)*, 11(3), 1-41. https://doi.org/10.1145/3374217

**Authors biography**

1. Gabriela Cioboata is a graduate student at the Faculty of Economics and Business Administration, Transilvania University of Brasov. Her main research interests are fake news detection and adversarial attacks against automated language models for sentiment analysis. Her research interest includes automated detection of such attacks and building more robust models. She has worked extensively on analysing model vulnerabilities to semantic inversion in financial contexts.
2. Teodor-Stefan Bildea, Ph.D, is an associate professor at the Faculty of Economics and Business Administration, Transilvania University of Brasov. His research interests are in the area of applications of data science, machine learning, and large language models to solving business problems.